\PassOptionsToPackage{table,dvipsnames}{xcolor}
\documentclass[10pt,twocolumn,letterpaper]{article}

\usepackage{iccv}      

\definecolor{iccvblue}{rgb}{0.21,0.49,0.74}
\usepackage[pagebackref,breaklinks,colorlinks,allcolors=iccvblue]{hyperref}

\usepackage{booktabs}
\usepackage{multirow}
\usepackage{amsmath} 
\usepackage{amssymb}
\usepackage{pifont}
\usepackage{subcaption} 
\usepackage{tabularx}
\usepackage{siunitx}

\def\paperID{16} 
\def\confName{ICCV}
\def\confYear{2025}

\title{SAGE: Segment-Aware Gloss-Free Encoding\\for Token-Efficient Sign Language Translation}

\author{Low Jian He\hspace{0.6cm} Ozge Mercanoglu Sincan\hspace{0.6cm} Richard Bowden \\[0.2cm]
CVSSP, University of Surrey, United Kingdom\\
\small{\texttt{\{jianhe.low, o.mercanoglusincan, r.bowden\}@surrey.ac.uk}}
}

\begin{document}
\maketitle

\begin{abstract}
Gloss-free Sign Language Translation (SLT) has advanced rapidly, achieving strong performances without relying on gloss annotations. However, these gains have often come with increased model complexity and high computational demands, raising concerns about scalability, especially as large-scale sign language datasets become more common.

We propose a segment-aware visual tokenization framework that leverages sign segmentation to convert continuous video into discrete, sign-informed visual tokens. This reduces input sequence length by up to 50\% compared to prior methods, resulting in up to 2.67× lower memory usage and better scalability on larger datasets. To bridge the visual and linguistic modalities, we introduce a token-to-token contrastive alignment objective, along with a dual-level supervision that aligns both language embeddings and intermediate hidden states. This improves fine-grained cross-modal alignment without relying on gloss-level supervision. Our approach notably exceeds the performance of state-of-the-art methods on the PHOENIX14T benchmark, while significantly reducing sequence length. Further experiments also demonstrate our improved performance over prior work under comparable sequence-lengths. Code release: \url{https://github.com/JianHe0628/SAGE}.

\end{abstract}    
\section{Introduction} \label{sec:intro}

Sign languages are the primary means of communication within Deaf communities and are natural human languages with rich linguistic structure and deep historical roots~\cite{peet1853elements}. They go beyond simple visual representations of speech, possessing their own phonological, morphological, and syntactic systems~\cite{Signvsspoken, DianeSignCompare}, distinct from spoken languages.

While recent advances in deep learning have transformed natural language understanding, most progress has been driven by text and speech modalities, supported by decades of research and large-scale resources. Sign language, in contrast, is inherently visual and multimodal, unfolding across space and time through coordinated hand gestures, facial expressions, and body movements \cite{koller2019weakly}. These multimodal signals interact in a continuous, fine-grained manner, making the modeling task especially difficult as no single modality or frame captures the full linguistic content.

\begin{figure}
    \centering
    \includegraphics[width=1.005\linewidth]{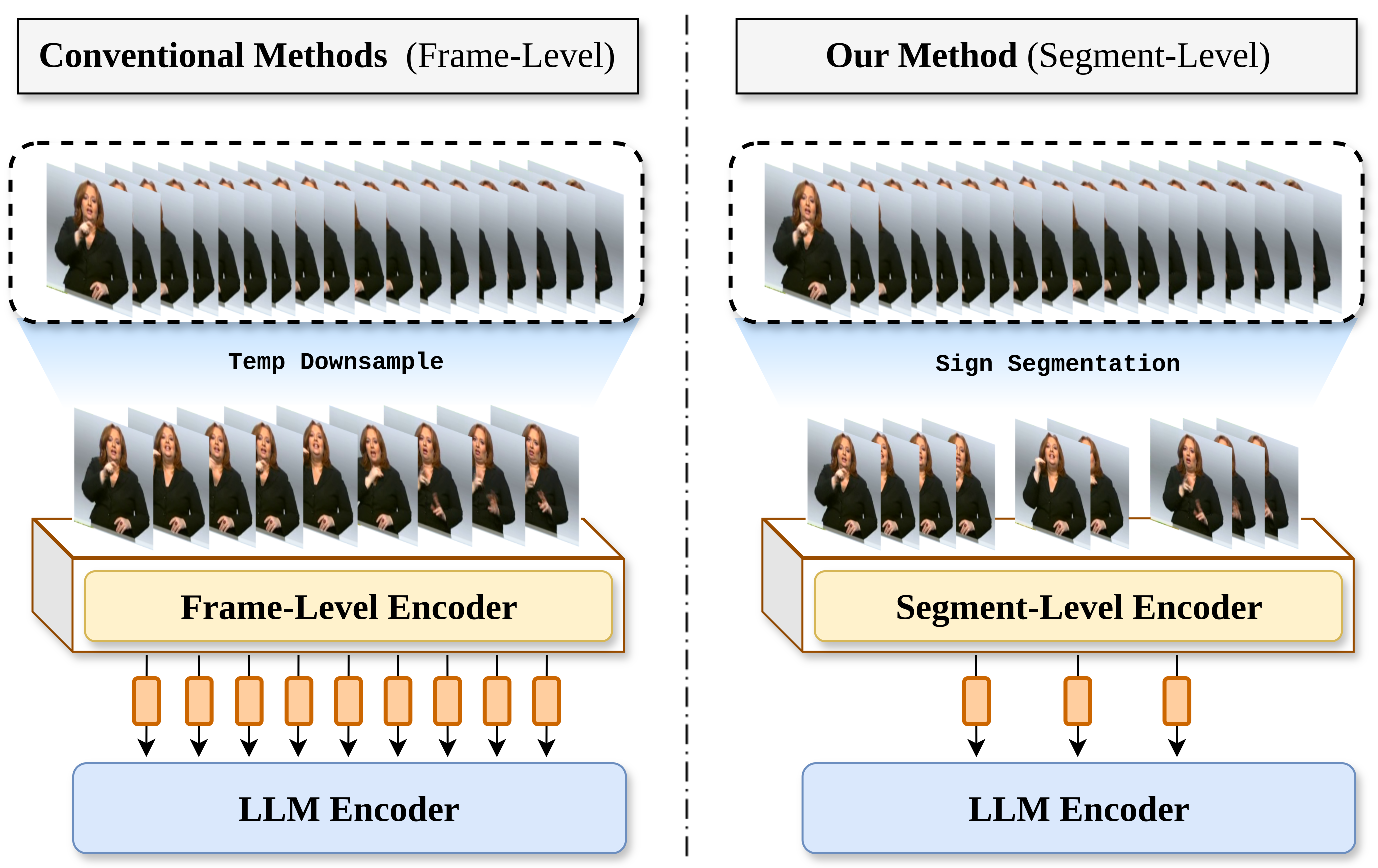}
    \caption{\textbf{Method Comparison.} Conventional approaches rely on temporal downsampling to reduce the number of video frames; however, this often still results in lengthy and redundant sequences being passed to the language encoder. In contrast, our method introduces a visual tokenization strategy that groups frames into discrete visual tokens, significantly reducing input sequence length.}
    \label{fig:teaser}
\end{figure}

Sign Language Translation (SLT), the task of translating continuous sign videos into spoken language sentences, is therefore notably challenging as it requires models to bridge complex spatiotemporal input and structured language output. A common strategy in prior work is to use gloss annotations as intermediate supervision~\cite{camgoz2020CSLR, chen2022simple, yin-read-2020-better, zhang2023sltunet, zhou2021improving}. Glosses are written representations of signs and thus provide explicit labels to guide models in recognizing signs before translation. While effective, gloss annotations are extremely costly~\cite{cardoso2021how2sign}, as they require frame-level labeling by expert linguists, creating a major bottleneck for scaling SLT systems.

To overcome this, recent efforts have shifted towards gloss-free SLT~\cite{camgoz2018neural,zhou2023gloss, wong2024sign2gpt, gong2024llms}, which bypasses intermediate supervision and trains directly on paired sign videos and spoken language sentences. These approaches often leverage contrastive learning, employing CLIP-style objectives~\cite{radford2021learning, zhou2023gloss} or frame-level contrastive losses~\cite{SignCL}, to align visual and textual embeddings. However, without explicit gloss labels, the supervision signal becomes weaker. To compensate, some works rely on large-scale pretraining~\cite{li2025uni, fish2025geo}, while others attempt to impose gloss-like alignment~\cite{wong2024sign2gpt}. In all cases, the absence of glosses leads to a heavier dependence on larger and more expressive architectures to learn the intricate cross-modal correspondences.

Additionally, the inherent fine-grained nature of sign language further exacerbates the modeling challenge. Sign language encodes meaning through subtle, continuous changes; thus, even slight variations can alter the semantic content, necessitating dense, frame-level processing to preserve linguistic fidelity. As recent SLT approaches have predominantly adopted Transformer-based visual encoders, full-resolution temporal encoding becomes computationally infeasible as the self-attention mechanism in Transformers scales quadratically with sequence length~\cite{lu2021soft}. To manage this, most approaches have introduced temporal reduction strategies, such as max pooling~\cite{zhou2023gloss}, uniform downsampling~\cite{wong2024sign2gpt, chen2024factorized}, and sliding window mechanisms~\cite{gong2024llms}. However, these methods overlook the linguistic structure of sign language, treating all frames as equally informative. This is problematic, as sign language exhibits variable sign duration and is sensitive to individual signing speeds~\cite{borstell2016distribution}. Thus, agnostic reduction risks discarding semantically critical frames, ultimately constraining a model's ability to capture the expressive richness of sign language.

Motivated by these challenges, we propose a \textit{segment-aware gloss-free encoding} framework for efficient SLT. As discussed, existing methods predominantly rely on frame-level operations; however, sign language segmentation, despite its inherent ability to group frames into coherent sign units, remains underutilized. Our implementation therefore explores leveraging a segmentation model \cite{he2025hands} as a visual tokenizer to localize meaningful sign units within continuous signing. Instead of processing dense sequences, we aggregate frames into semantically coherent segments and encode each segment as a single visual token. This significantly reduces sequence length, alleviates the computational burden associated with Transformer-based encoders, and more faithfully captures the linguistic structure of sign language. As shown in Fig.~\ref{fig:teaser}, our method yields significantly fewer tokens than frame-based downsampling. To align visual tokens with language, we further introduce a \textit{token-level contrastive alignment objective} that encourages fine-grained correspondence between visual and text tokens. In summary, our main contributions are as follows:
\begin{itemize}
    \item We propose the first gloss-free SLT framework that incorporates sign language segmentation to produce semantically aligned, segment-based visual tokens.
    \item Our approach achieves substantial temporal reduction, significantly lowering token count and Transformer computation while maintaining translation quality.
    \item We introduce a token-level contrastive alignment objective that enhances the correspondence between visual segments and language representations.
    \item We conduct ablations and experiments on PHOENIX14T, demonstrating that our method achieves translation performances exceeding prior state-of-the-art.
\end{itemize}
\section{Related Work} \label{sec:relatedwork}

Sign Language Understanding (SLU) spans a wide range of tasks aimed at interpreting signed communication from visual input. Isolated Sign Language Recognition (ISLR) focuses on classifying individual signs from short, pre-segmented video clips~\cite{li2020ISLR, albanie2020bsl1k, ISLR2021AAAI, jiang2021skeletonISLR, wong2023ISLR, zuo2023ISLR}, while Continuous Sign Language Recognition (CSLR) targets the recognition of co-articulated sign sequences within continuous video streams. While both tasks are crucial in SLU, CSLR better reflects the natural temporal and linguistic structure of sign language and has been advanced considerably by gloss-annotated datasets such as PHOENIX14~\cite{koller2015continuous} and CSL-Daily~\cite{zhou2021improving}. However, the lack of frame-level alignment in these datasets has led to the widespread adoption of Connectionist Temporal Classification (CTC) as a weakly supervised learning approach~\cite{hu2023CSLR, min2021CSLR, niu2020CSLR, pu2019CSLR, zuo2022C2SLR}.

Despite the progress in CSLR, the broader objective of SLU lies in Sign Language Translation (SLT), the task of generating spoken-language translations directly from sign language videos. Unlike CSLR, which outputs gloss sequences, SLT aims to produce coherent natural language sentences, thus offering a more accessible communication channel for both Deaf and hearing communities. Camgoz et al.~\cite{camgoz2018neural} first formalized SLT as a sequence-to-sequence learning problem, establishing a foundation for subsequent advancements. SLT methods can broadly be categorized into \textit{gloss-based} and \textit{gloss-free} approaches.

\textbf{Gloss-based approaches} rely on gloss annotations either as an explicit intermediate representation or as auxiliary supervision. For example, SLRT~\cite{camgoz2020CSLR} employed a Transformer-based encoder-decoder architecture conditioned on gloss sequences. SignBT~\cite{zhou2021improving} improved on this pipeline by using back-translation, a concept adapted from machine translation. A multi-modal transfer learning-based (MMTLB) approach ~\cite{chen2022simple} was then proposed by pretraining on general-domain datasets, while CV-SLT~\cite{zhao2024conditional} enhanced this by integrating a conditional variational autoencoder to improve cross-modal generalization. Other efforts, such as TS-SLT~\cite{twostream}, also explored two-stream architectures that leverage both raw video and keypoint sequences.

In general, gloss-based methods significantly outperform gloss-free approaches due to the finer-grained supervision; however, they face the critical limitation of being dependent on gloss annotations. This requirement poses a scalability challenge, as gloss annotations are labor-intensive and not consistently available across datasets and sign languages.

 \begin{figure*}[t]
    \centering
    \includegraphics[width=1\linewidth]{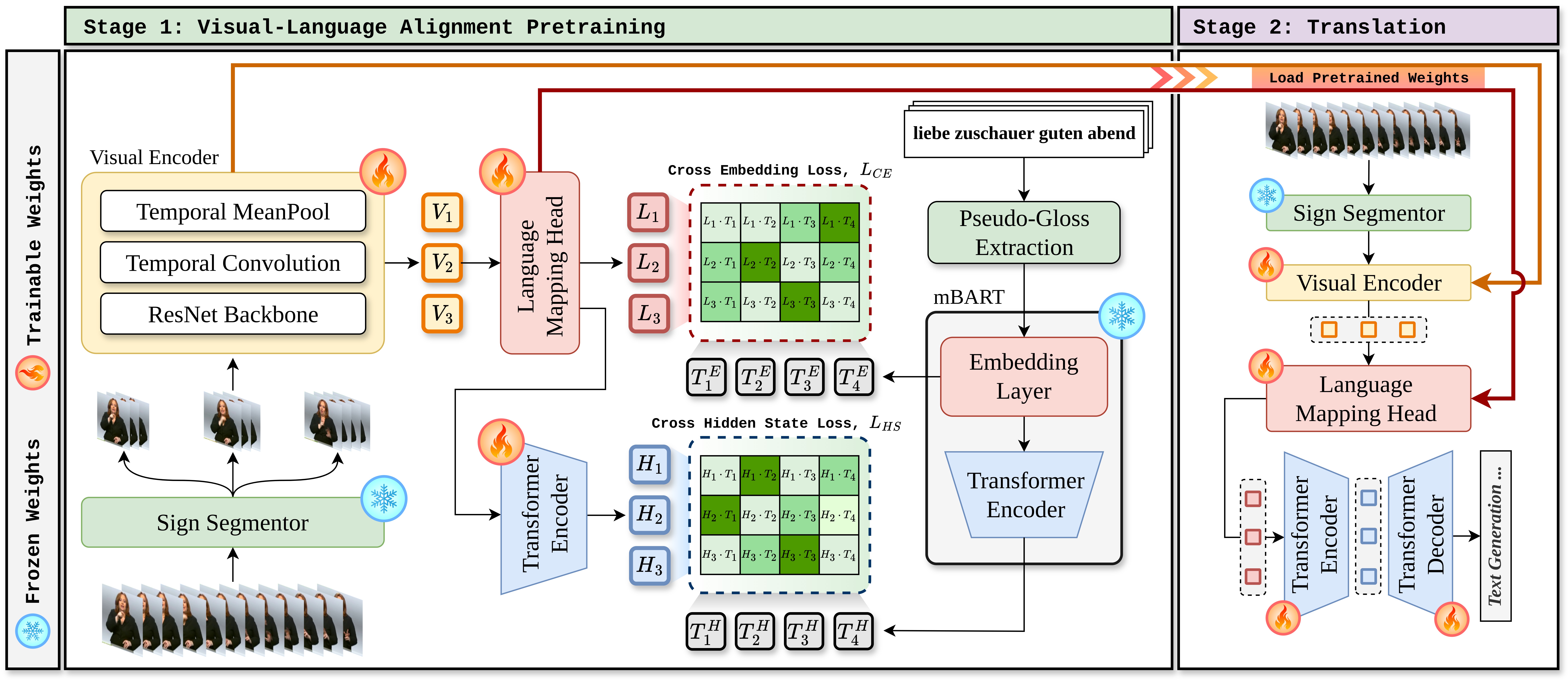}
    \caption{\textbf{Architecture Overview.} Our method consists of two stages. (i) In the first stage, a sign segmentor tokenizes continuous sign videos, and a visual encoder maps each segment into an embedding. We then apply contrastive pretraining to align the visual representations with the textual embeddings from a language encoder, encouraging semantically rich and discriminative features. (ii) In the second stage, the visual tokenizer and pretrained encoder are integrated into a sequence-to-sequence model and fine-tuned for sign language translation.}
    \label{fig:overall architecture}
\end{figure*}

\textbf{Gloss-free methods}, in contrast, bypass gloss annotations entirely and learn directly from video and spoken language pairs. While they typically underperform gloss-based systems due to weaker supervision, they offer a scalable alternative by removing the need for gloss annotations. This approach is becoming increasingly viable with the emergence of large-scale gloss-free datasets such as YTSL-25~\cite{tanzer2024youtube} and CSL-News~\cite{li2025uni}. As a result, the field has rapidly pivoted towards gloss-free SLT, prioritizing architecture designs and training strategies that can compensate for the absence of gloss through alternative learning signals.

For instance, GASLT~\cite{yin2023gloss} introduced a gloss attention mechanism to inject gloss-like inductive bias into the learning process. GFSLT~\cite{zhou2023gloss}, leveraged a CLIP-based~\cite{radford2021learning} architecture, and was pretrained on video-sentence pairs via cross-modal alignment. Sign2GPT~\cite{wong2024sign2gpt} introduced pseudo-gloss generation during pretraining to recover some level of intermediate supervision. SignCL~\cite{SignCL} proposed a contrastive learning framework by treating temporally adjacent frames as semantically coherent units. FLa-LLM~\cite{chen2024factorized} incorporated a lightweight Transformer, and proposed pretraining with a language modeling objective. Lastly, SignLLM~\cite{gong2024llms} explored a discrete codebook to learn gloss-like token representations in a self-supervised manner.

While all aforementioned methods showcase a wide range of strategies, most adopt Transformer-based encoders to process video frames, resulting in increased computation. In this work, we thus address these limitations by proposing a more efficient gloss-free SLT framework that maintains strong performances while reducing complexity.

\section{Methodology}
Our proposed framework for SLT addresses two key limitations in existing approaches: the use of linguistically-agnostic temporal sampling, which fails to respect the semantic boundaries of signs, and the high memory complexity of current Transformer-based models, which hinders scalability to larger datasets. To this end, we introduce a two-stage SLT architecture (in Fig. \ref{fig:overall architecture}), that uses a novel visual tokenizer to segment continuous sign videos into discrete units, allowing for more memory efficient translation.

\subsection{Stage 1: Visual-Language Pretraining} \label{stage1}
In the first stage, our goal is to pretrain a visual encoder such that it maps segmented sign units into semantically meaningful embeddings aligned with word-level linguistic representations. Instead of aligning directly to spoken sentences, which may contain syntactic noise, we adopt a pseudo-gloss formulation that provides a more compact and semantically focused target for alignment. We discuss our overall pretraining architecture further in the following subsections, including the segment-aware visual tokenization, pseudo-gloss extraction pipeline and contrastive learning objective.

\subsubsection{Segment-Aware Visual Tokenization} \label{sec:vis tokenization}

Given a continuous sign language video consisting of $T$ frames, denoted as $\mathbf{X} = \{x_1, x_2, \dots, x_T\}$, our goal is to segment the video into $N$ temporally localized sign units using motion and appearance cues. This results in a set of segments $\mathbf{X}' = \{\mathbf{S}_1, \mathbf{S}_2, \dots, \mathbf{S}_N\}$, where each $\mathbf{S}_i \in \mathbb{R}^{n_i \times c}$ corresponds to a contiguous sequence of $n_i$ frames with $c$-dimensional image features. These segments are subsequently passed into a visual encoder (Section \ref{sec:spatiotemp enc}) to produce a sequence of sign tokens $\mathbf{t} = \{\mathbf{t}_1, \mathbf{t}_2, \dots, \mathbf{t}_N\}$, where each $\mathbf{t}_i \in \mathbb{R}^c$ captures the motion dynamics and fine-grained spatial information of a localized sign unit.

To perform sign segmentation, we utilize the recently proposed \textit{Hands-On} model~\cite{he2025hands}, which segments signing sequences based on input features from HaMeR hand representations~\cite{pavlakos2024reconstructing} and 3D body pose trajectories~\cite{ivashechkin2023improving}. In our pipeline, the Hands-On model is used in a frozen state, without task-specific fine-tuning, relying on its learned priors over hand shape and upper-body motion to generalize across datasets. For additional details on the segmentation model, runtime analysis, and visual tokenization implementation, please refer to the supplementary materials.

\subsubsection{Language Encoder} \label{sec:language enc}

\paragraph{Pseudo-Gloss Extraction:} Before extracting textual embeddings, we first convert the spoken sentence into a pseudo-gloss sequence that serves as a compact and semantically grounded alignment target. Given a spoken sentence $\mathbf{S} = \{w_1, w_2, \dots, w_L\}$ of length $L$, we derive a pseudo-gloss sequence $\mathbf{G} = \{g_1, g_2, \dots, g_M\}$, where $M < L$, by filtering for content-bearing words. Following~\cite{wong2024sign2gpt}, we apply a part-of-speech (POS) tagging pipeline and retain only tokens tagged as nouns, verbs, adjectives, numerals, adverbs, pronouns, or proper nouns. This process removes syntactic noise while preserving the semantic core of the sentence, yielding a gloss-like abstraction that is better aligned with the compositional structure of signing.

\paragraph{Textual Representation.}  
To obtain linguistic features for alignment, we use \textit{mBART-large-50}~\cite{tang2020multilingual} as our language encoder. Thus, given a sequence of pseudo-glosses $\mathbf{G}$, the model first tokenizes them into a sequence of subword tokens $\{t_1, t_2, \dots, t_M\}$. Each token $t_i$ is then embedded via an input embedding matrix $E \in \mathbb{R}^{d \times {1024}}$, with $d$ as the vocabulary size, yielding:
\begin{equation}
\mathbf{T}^{E} = \{\mathbf{e}_i = E[t_i] \in \mathbb{R}^{1024} \mid i = 1, \dots, M\},   
\end{equation}
where token-level embeddings $\mathbf{T}^{E}$ capture lexical identity but lack contextual semantics, and serve as type-level linguistic representations in our contrastive alignment objective. To capture contextual dependencies between tokens, we pass $\mathbf{T}^{E}$ through the mBART Encoder $ME$ to obtain contextualized hidden states:
\begin{equation}
\mathbf{T}^{H} = ME(\mathbf{T}^{E}) = \{\mathbf{h}_1, \mathbf{h}_2, \dots, \mathbf{h}_M\},
\end{equation}

\noindent where each embedding $\mathbf{h}_i \in \mathbb{R}^{1024}$ encodes context-aware semantics via self-attention over the entire input sequence.

Unlike prior approaches that align visual features solely with the final hidden state outputs of a language encoder, we leverage both $\mathbf{T}^{E}$ and $\mathbf{T}^{H}$ in our contrastive objective. This dual representation enables supervision at both lexical and contextual levels, capturing stronger fine-grained and compositional structure inherent to sign language.

\subsubsection{Visual Encoder} \label{sec:spatiotemp enc}

To represent the visual modality at the sign segment level, we construct a hierarchical encoder that captures both spatial and temporal structure across consecutive video frames. Each sign segment $S = \{s_1, s_2, \dots, s_n\}$, comprising of $n$ frames, is first processed by a ResNet-34 to extract frame-wise spatial features. This results in a sequence of embeddings $V = \{v_1, v_2, \dots, v_n\}$, where $v_i \in \mathbb{R}^{512}$. Meanwhile, local temporal dependencies within the segment are captured via a lightweight temporal encoder consisting of a 1D convolutional layer (kernel size $k = 5$), followed by Batch Normalization and ReLU activation. In parallel, the temporal encoder also up-projects the features, resulting in a spatiotemporal representation of $V^{\text{temp}} \in \mathbb{R}^{(n - 4) \times 1024}$.

We then apply average pooling across the temporal dimension to obtain a singular token representation $V^{\text{tok}} \in \mathbb{R}^{1024}$ that summarizes both spatial and temporal dynamics of the sign segment. Finally, to capture contextual relationships across segments, we introduce a Transformer-based encoder $\text{VE}$ that operates over the sequence of tokens and outputs hidden states \( \mathbf{H} \in \mathbb{R}^{1024} \). Since the token sequence is much shorter than the original frame sequence, our design enables efficient global reasoning while maintaining fine-grained semantic structure on the full signing sequence.

\subsubsection{Vision-to-Language Mapper}
To bridge the gap between visual and linguistic modalities, we introduce a \textit{Visual-to-Language Mapper} that projects visual sign embeddings into the token embedding space of a pretrained language model. Given a visual token representation \(V^{\text{tok}} \in \mathbb{R}^{1024}\), the mapper \(f_{\mathrm{map}}\) produces a corresponding language embedding \(\mathbf{L} = f_{\mathrm{map}}(V^{\text{tok}}) \in \mathbb{R}^{1024}\). The function \(f_{\mathrm{map}}\) is implemented as a stack of three feedforward blocks, each consisting of Layer Normalization, a linear projection, GeLU activation, and Dropout, following Transformer-inspired architectures~\cite{devlin2019bert}. We supervise the mapper to align \(\mathbf{L}\) with the token embeddings \(\mathbf{T}^{E}\) from the language encoder, ensuring the mapped representations reside in a semantically meaningful space.

\subsubsection{Contrastive Learning Objective}

\begin{figure}[t]
    \centering
    \label{fig: contrastive compare}
    \includegraphics[width=1\linewidth]{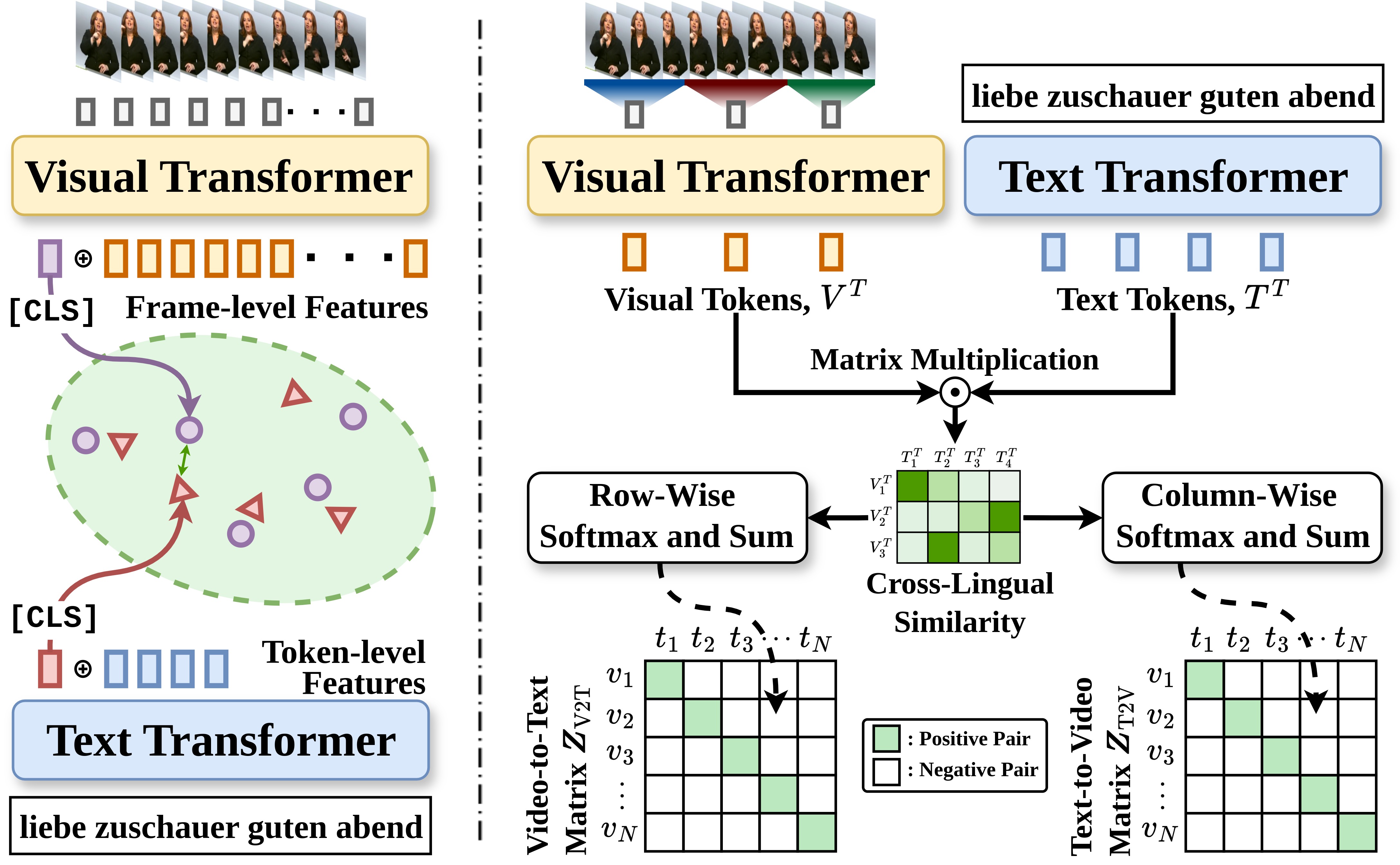}
    \caption{\textbf{Contrastive Learning Comparison.} Prior pretraining methods adopt CLIP-style objectives to align video and text globally via summarized [CLS] tokens. In contrast, our approach performs fine-grained cross-lingual alignment by computing token-level similarities across all the video-text pairs within a batch.}
    \label{fig:enter-label}
\end{figure}

To enable fine-grained alignment between visual and textual tokens during pretraining, we require an objective that captures cross-modal correspondences without relying on strict ordering. As our pseudo-glosses follow spoken language order rather than the temporal progression of the video, monotonic alignment strategies such as CTC are unsuitable.

Given these limitations, contrastive learning offers a more flexible alternative for cross-modal supervision. However, widely adopted CLIP-style methods~\cite{radford2021learning} typically operate at the global level, compressing entire videos and sentences into single [CLS] embeddings. While proven effective for SLT~\cite{zhou2023gloss}, such global representations discard the rich internal structure of both modalities, making them less suited for the token-level alignment our framework requires.

To address this, we adopt a more granular supervision objective based on the \textit{Cross-Lingual Contrastive Learning (CLCL)} loss~\cite{cheng2023cico}, originally introduced to learn a joint sign video and spoken-language embedding space while capturing fine-grained sign-to-word mappings. Unlike the original implementation, which uses per-frame I3D features as sign representations, we apply CLCL at the token level by aligning segment-level visual tokens to language tokens.

In addition to token-level alignment, we also follow CLCL by computing a global video-to-text similarity matrix $\boldsymbol{Z}_{\text{V2T}} \in \mathbb{R}^{N \times N}$ for each mini-batch, where $\boldsymbol{Z}_{\text{V2T}}^{(i,j)}$ denotes the similarity between the $i$-th video and the $j$-th sentence. We then apply the InfoNCE loss~\cite{gutmann2010noise}, formulated as:

\begin{equation}
  \label{loss:NCE}
  \begin{split}
\mathcal{L}_{\text{V2T}}= & -\frac{1}{2N}\sum_{i=1}^{N} \log \frac{\exp(\boldsymbol{Z}_{\text{V2T}}^{(i,i)}/\tau)}{\sum_{j=1}^{N} \exp(\boldsymbol{Z}_{\text{V2T}}^{(i,j)}/\tau)} \\
& -\frac{1}{2N}\sum_{j=1}^{N} \log \frac{\exp(\boldsymbol{Z}_{\text{V2T}}^{(j,j)}/\tau)}{\sum_{i=1}^{N} \exp(\boldsymbol{Z}_{\text{V2T}}^{(i,j)}/\tau)},
  \end{split}
\end{equation}

\noindent where $\tau$ is a learnable temperature parameter. We similarly compute the text-to-video loss $\mathcal{L}_{\text{T2V}}$ by using the similarity matrix $\boldsymbol{Z}_{\text{T2V}}$, and define the final loss as a weighted sum:

\begin{equation}
  \label{loss:final}
  \mathcal{L}_{\text{CLCL}} = \alpha \mathcal{L}_{\text{V2T}} + (1 - \alpha) \mathcal{L}_{\text{T2V}},
\end{equation}

\noindent where $\alpha \in [0,1]$ controls the contribution of each alignment direction.
Furthermore, as described in Section~\ref{sec:language enc}, we enhance the contrastive supervision by applying the CLCL loss to both the input embeddings $\mathbf{T}^{E}$ and the contextualized hidden states $\mathbf{T}^{H}$ of the language encoder. As the two contrastive objectives are applied separately we refer to them as the \textit{Cross-Embedding Loss}  $\mathcal{L}_{\text{CE}}$ and \textit{Cross-Hidden-State Loss} $\mathcal{L}_{\text{HS}}$, respectively, as illustrated in Fig.~\ref{fig:overall architecture}. The final pretraining objective is then formulated as a weighted sum of the two, controlled by a hyperparameter $\beta \in [0,1]$ which is tuned through ablation experiments in Table~\ref{tab:LossBalancing}:

\begin{equation}
\label{eq:final loss}
\mathcal{L}_{\text{total}} = \beta \, \mathcal{L}_{\text{CE}} + (1 - \beta) \, \mathcal{L}_{\text{HS}},
\end{equation}


\subsection{Stage 2: Sign Language Translation}

In the second stage, our objective is to fine-tune our model for the downstream task of SLT. We begin by initializing the visual encoder and the Visual-to-Language Mapper with pretrained weights from the alignment stage (see Fig.~\ref{fig:overall architecture}). Since this stage already aligns visual embeddings with linguistic representations, it provides a strong language-aware prior for downstream translation. To complete the architecture, we attach the full \textit{mBART-large-50} encoder-decoder model, where the mBART encoder consumes the mapped visual tokens and the decoder autoregressively generates the target spoken sentence. This setup allows for end-to-end translation of sign videos into spoken language sentences.

To ensure consistency with the pretraining pipeline and to maintain computational efficiency, we  also retain the segment-based tokenizer to produce compact visual tokens. Compared to prior frame-based methods, this approach significantly reduces sequence length while preserving semantic granularity in SLT. Thus, given an input sign sequence \(\mathbf{X} = \{x_1, x_2, \dots, x_T\}\) of \(T\) frames, we first obtain a sequence of visual tokens \(\mathbf{V} = \{v_1, \dots, v_N\} \in \mathbb{R}^{N \times 1024}\) via the visual encoder. These are then projected into the linguistic space through the mapper, yielding \(\mathbf{L} = \{l_1, \dots, l_N\} \in \mathbb{R}^{N \times 1024}\), which are then processed by the mBART encoder to produce hidden states \(\mathbf{H}\). The decoder subsequently generates the translated output sentence \(\mathbf{S}' = \{s'_1, \dots, s'_M\}\) token-by-token, and the entire model is optimized using the standard language modeling objective:
\begin{equation}
\mathcal{L}_{\text{LM}} = - \sum_{m=1}^{M} \log P(s'_m \mid s'_{<m}, \mathbf{X}),
\end{equation}

\noindent where \(s'_m\) denotes the \(m\)-th predicted token and \(s'_{<m}\) its preceding tokens. This objective thus trains the decoder to generate coherent translations conditioned on the visual input.

\section{Experiments}

\subsection{Datasets and Evaluation Metrics}

\paragraph{Dataset.} 
To ensure comparability with prior work, we evaluate our model on the widely adopted \textit{RWTH-PHOENIX-Weather 2014T (PHOENIX14T)} benchmark~\cite{camgoz2018neural}. This dataset consists of German Sign Language (DGS) videos extracted from weather forecast broadcasts, and is annotated with both gloss-level transcriptions as well as corresponding spoken German sentences. It consists of 8,257 video clips, amounting to approximately 11 hours of footage. The standard splits include 7,096 samples for training, 519 for validation, and 642 for testing.

\vspace{-1em}

\paragraph{Evaluation Metrics.}
Following established SLT evaluation protocols, we report performance using \textit{Bilingual Evaluation Understudy (BLEU)}~\cite{papineni2002bleu} and \textit{Recall-Oriented Understudy for Gisting Evaluation (ROUGE-L)}~\cite{lin2004rouge}. BLEU measures the n-gram overlap between predicted and reference translations, capturing precision over different granularities. We report BLEU scores from unigram to 4-gram (BLEU-1 to BLEU-4). ROUGE-L evaluates translation quality based on the longest common subsequence, and considers both precision and recall. We follow the standard practice of reporting the F1-score variant of ROUGE-L.

\subsection{Implementation Details}

\paragraph{Stage 1: Contrastive Pretraining.} 
We train the visual encoder and the Vision-to-Language Mapper using a batch size of 16 and stochastic gradient descent (SGD) with an initial learning rate of 0.03, momentum of 0.9, and a cosine annealing learning rate schedule. Gradient clipping is applied with a threshold of 1.0, and a dropout rate of 0.1 is used in the mapper to mitigate overfitting. Training is performed for 80 epochs on a single NVIDIA A100 GPU.

To enhance generalization, we apply slight data augmentations following~\cite{zhou2023gloss}, including random rotation, color jitter, and brightness adjustments. All video frames are resized to \(256 \times 256\), followed by center cropping to \(224 \times 224\) to ensure input consistency. In line with prior work, we also trim the mBART tokenizer's embedding matrix to retain only vocabulary tokens present in the training set.

\vspace{-1em}
\paragraph{Stage 2: Downstream Translation.} 
When fine-tuning for SLT, we use a batch size of 8 and optimize with SGD. The learning rate is set to 0.004, momentum to 0.9, and weight decay to 0.001. Gradient clipping is set to 5.0, and dropout remains at 0.1. Training is again conducted for 80 epochs on a single NVIDIA A100. Similar data augmentations are applied and the model is trained using cross-entropy with label smoothing of 0.2. During inference, we configure the mBART decoder with a maximum token length of 150 and set number of beams to 4 for decoding.

\begin{table}[t]
\centering
\small
\begin{tabular}{lccccc}
\toprule
\textbf{Method} & \textbf{B1} & \textbf{B2} & \textbf{B3} & \textbf{B4} & \textbf{R-L} \\
\midrule
\rowcolor{gray!20}
\multicolumn{6}{c}{\textbf{Gloss-Based}} \\
\midrule
SL-T \cite{camgoz2020CSLR}         & 46.61 & 33.73 & 26.19 & 21.32 & --    \\
SignBT \cite{zhou2021improving}   & 50.80 & 37.75 & 29.72 & 24.32 & 49.54 \\
MMTLB \cite{chen2022simple}       & 53.97 & 41.75 & 33.84 & 28.39 & 52.65 \\
SLTUNET \cite{zhang2023sltunet}   & 52.92 & 41.76 & 33.99 & 28.47 & 52.11 \\
TS-SLT \cite{twostream}           & 54.90 & 42.43 & 34.46 & 28.95 & 53.48 \\
CV-SLT \cite{zhao2024conditional} & 54.88 & 42.68 & 34.79 & 29.27 & 54.33 \\
\midrule
\rowcolor{gray!20}
\multicolumn{6}{c}{\textbf{Gloss-Free}} \\
\midrule
NSLT \cite{camgoz2018neural}       & 29.86 & 17.52 & 11.96 & 9.00  & 30.70 \\
TSPNet \cite{li2020tspnet}         & 36.10 & 23.12 & 16.88 & 13.41 & 34.96 \\
CSGCR \cite{CSGCR}                & 36.71 & 25.40 & 18.86 & 15.18 & 38.85 \\
GASLT \cite{yin2023gloss}          & 39.07 & 26.74 & 21.86 & 15.74 & 39.86 \\
GFSLT-VLP \cite{zhou2023gloss}    & 43.71 & 33.18 & 26.11 & 21.44 & 42.49 \\
Sign2GPT \cite{wong2024sign2gpt}   & 49.54 & 35.96 & 28.83 & 22.52 & \textbf{48.90} \\
FLa-LLM \cite{chen2024factorized}  & 46.29 & 35.33 & 28.03 & 23.09 & 45.27 \\
SignLLM \cite{gong2024llms}        & 45.21 & 34.78 & 28.05 & 23.40 & 44.49 \\
\midrule
\textbf{SAGE (Ours)}                              & \textbf{49.69} & \textbf{37.01} & \textbf{29.21} & \textbf{24.10} & 48.86 \\
\bottomrule
\end{tabular}
\caption{\textbf{Sign Language Translation results on PHOENIX14T (test set).} We compare against recent SOTA gloss-free SLT models. \textbf{Bold} indicates best performance among gloss-free methods.}

\label{tab:sotacompare}
\end{table}

\subsection{Comparisons to State-of-the-Art Methods}
To assess our token-efficient approach, we conduct extensive comparisons against recent state-of-the-art (SOTA) methods. We evaluate both the maximum translation performance (Section \ref{sec:max trans}) and the computational efficiency (Section \ref{sec:efficiencycompare}), including comparisons of total memory usage and performance under similar token reduction settings.

\subsubsection{Maximum Translation Performance Comparisons} \label{sec:max trans}
In Table~\ref{tab:sotacompare}, we compare our model against recent state-of-the-art (SOTA) methods on the PHOENIX14T test set. Our approach achieves the highest performance across all BLEU scores, surpassing previous methods, including a +0.7 increase in BLEU-4 (24.10 vs. 23.40). The gains in BLEU-1 to BLEU-3 suggest that our token-level supervision is particularly effective at capturing fine-grained, word-level correspondences. This is likely due to our design, where sign videos are represented as discrete tokens and directly aligned with text tokens during training, successfully reducing the gap between visual input and linguistic output at the word level. Beyond token-level matches, our improvement in BLEU-4 further indicates strong performance at the sentence level. This suggests that the visual tokens not only provide accurate local alignments, but also serve as robust priors for the language decoder to generate coherent and fluent translations. Additionally, our ROUGE score closely matches that of Sign2GPT, highlighting that our model preserves strong semantic similarity to the reference translations. Importantly, we achieve these results with a significantly more efficient and lightweight architecture, demonstrating the potential of visual tokenization as an effective and scalable representation for sign language translation.

\definecolor{darkgreen}{RGB}{0,128,0} 
\subsubsection{Computational Efficiency Comparisons} \label{sec:efficiencycompare}
\begin{table}[t]
\centering
\small
\begin{tabular}{lcccccc}
\toprule
\textbf{Method} & \textbf{MP} & \textbf{SD} & \textbf{SW} & \textbf{UD} & \textbf{Ratio} & \textbf{B4} \\ 
\midrule
\multirow{2}{*}{GFSLT$^\dagger$ \cite{zhou2023gloss}} 
  & \textcolor{darkgreen}{\ding{51}} & \textcolor{red}{\ding{55}} & \textcolor{red}{\ding{55}} & --     & 0.25  & 21.44 \\
  & \textcolor{darkgreen}{\ding{51}} & \textcolor{red}{\ding{55}} & \textcolor{red}{\ding{55}} & T/2   & \underline{0.125} & \underline{19.25} \\
\addlinespace
Sign2GPT \cite{wong2024sign2gpt}
  & \textcolor{red}{\ding{55}} & \textcolor{darkgreen}{\ding{51}} & \textcolor{red}{\ding{55}} & T/2     & 0.25  & 22.52 \\
\addlinespace
\multirow{2}{*}{FLa-LLM \cite{chen2024factorized}} 
  & \textcolor{red}{\ding{55}} & \textcolor{red}{\ding{55}} & \textcolor{red}{\ding{55}} & T/4   & 0.25  & 23.09 \\
  & \textcolor{red}{\ding{55}} & \textcolor{red}{\ding{55}} & \textcolor{red}{\ding{55}} & T/8 & \underline{0.125} & \underline{20.02} \\
\addlinespace
SignLLM \cite{gong2024llms}
  & \textcolor{red}{\ding{55}} &\textcolor{red}{\ding{55}} & \textcolor{darkgreen}{\ding{51}} & --     & 0.25  & 23.40 \\

\arrayrulecolor{gray} \midrule
\textbf{SAGE (Ours)} 
  & \multicolumn{4}{l}{\textbf{Visual Tokenization}}       &\textbf{0.129} & \textbf{24.10} \\
\arrayrulecolor{black} \bottomrule
\end{tabular}
\caption{\textbf{Comparison of Temporal Reduction Strategies.} We compare our Visual Tokenization approach with Max Pooling (MP), Strided Downsampling (SD), Sliding Window (SW), and Uniform Downsampling (UD). Reported reduction ratios provide context on computational efficiency, with those comparable to ours \underline{underlined} for emphasis. $\dagger$ indicates results reproduced by us.}
\label{tab:CompareRatio}
\end{table}

\begin{table}[t]
\centering
\begin{tabular}{lrr}
\toprule
\textbf{Method} & \textbf{VRAM (Total)} & \textbf{Base GPU Req.} \\ \midrule
GFSLT~\cite{zhou2023gloss}     & 80 GB    & 1 × A100      \\
Sign2GPT~\cite{wong2024sign2gpt}  & 160 GB   & 2 × A100      \\
SignLLM~\cite{gong2024llms}    & 96 GB    & 4 × A5000     \\
\arrayrulecolor{gray} \midrule
\textbf{SAGE (Ours)}           & $\sim$ \textbf{60 GB} & $\sim$ \textbf{3 × RTX 3090}  \\ \arrayrulecolor{black} \bottomrule
\end{tabular}
\caption{\textbf{Peak GPU Memory Usage During Training.} We report the peak VRAM consumption and minimum GPU required based on the training setups used by each method to achieve their respective reported performances. For our method, the reported values are approximate due to minor variation in input sequence lengths.}
\label{tab:vramcompare}
\end{table}

To further contextualize our contribution, we evaluate how prior SOTA models perform under comparable sequence reduction constraints. Existing approaches typically reduce input frame lengths using fixed downsampling strategies, such as strided sampling, sliding windows, or pooling; and they are often applied at a factor of 4 (i.e., a reduction ratio of ~0.25). In contrast, our method employs a visual tokenization strategy that maps input frames into a compact sequence of semantic tokens, achieving a significantly higher compression rate with a reduction ratio of 0.129, nearly halving the sequence length compared to typical methods.

To assess the impact of such compression on translation quality, we evaluate how prior models perform under similar levels of temporal reduction. Specifically, we retrain GFSLT using T/2 frame downsampling and include the reported T/8 results for FLa-LLM \cite{chen2024factorized} from their original work. While such downsampling strategies are commonly used, we acknowledge that they may substantially degrade the fidelity of the input signal. However, these strategies represent some of the only practical options currently available in SLT for reducing sequence length, highlighting the need for more semantically informed compression strategies. As shown in Table~\ref{tab:CompareRatio}, these models exhibit notable performance drops under increased compression; with FLa-LLM, for instance, falling to a BLEU-4 of 20.02 (-3.07). In contrast, our approach offers strong translation performance (BLEU-4: 24.10) despite the compact sequence length.

Table~\ref{tab:vramcompare} compares the peak GPU memory consumption of our model against recent SOTA. This comparison is especially important given the reliance of SLT architectures on contrastive learning, which is highly sensitive to the \textit{effective batch size} during training. Larger batch sizes allow for a greater number of negative pairs; however, achieving such large batches is often constrained by GPU memory. Fortunately, our method requires significantly less peak GPU memory compared to prior SOTA, using up to 2.67× less VRAM than Sign2GPT, resulting in more efficient training and improved scalability to larger-scale datasets.

\subsection{Ablation Study}
To assess the effectiveness of various architectural and training design choices in our model, we conduct extensive ablation studies on the \textsc{PHOENIX14T} development set, focusing on the BLEU-4 and ROUGE-L scores.

\vspace{-1em}

\paragraph{Pretraining Loss Configuration.}
We first investigate the effect of different contrastive loss formulations and the role of our proposed dual-level supervision. As shown in Table~\ref{tab:LossConfig}, using the fine-grained CLCL loss consistently outperforms CLIP-style global alignment, demonstrating the importance of token-level contrastive supervision in closing the visual-language gap. Furthermore, results also indicate that dual-level supervision within the language encoder provides a stronger alignment signal, as it enhanced model performances regardless of whether CLIP or CLCL was used.

\vspace{-1em}

\paragraph{Loss Balancing.}
In Table~\ref{tab:LossBalancing}, we then ablate the weighting coefficient $\beta$ between the Cross-Embedding and Cross-Hidden-State losses as defined in Eq. \ref{eq:final loss}. Varying $\beta$ from 0 to 0.8, we find that $\beta = 0.6$, which places more emphasis on the embedding-level loss, achieves the highest BLEU-4 score, while maintaining a ROUGE-L score nearly on par with the best. These results suggest that prioritizing alignment at the earlier language representation stage provides a stronger supervisory signal for downstream translation.

\begin{table}[t]
    \begin{subtable}[b]{0.6\columnwidth}
        \centering
        \resizebox{\linewidth}{!}{ 
            \begin{tabular}{l|c|cc}
            \toprule
            \textbf{Loss} & \textbf{DS} & \textbf{BLEU4} & \textbf{ROUGE} \\ \midrule
            CLIP & \textcolor{red}{\ding{55}} &  21.93  &    47.04  \\
            CLIP & \textcolor{darkgreen}{\ding{51}}  & 22.30   &   47.88   \\
            CLCL & \textcolor{red}{\ding{55}} & 23.25   &  48.10 \\
            \rowcolor{gray!20}
            \textbf{CLCL}   &  \textcolor{darkgreen}{\ding{51}}  & \textbf{23.81} &  \textbf{49.08}  \\ \bottomrule
            \end{tabular}
        }
        \caption{Different Loss Configurations}
\label{tab:LossConfig}
    \end{subtable}
    \hfill 
    \begin{subtable}[b]{0.385\columnwidth}
        \centering
        \resizebox{\linewidth}{!}{
            \begin{tabular}{c|cc}
            \toprule
                \textbf{$\boldsymbol{\beta}$} & \textbf{BLEU4} & \textbf{ROUGE} \\ \midrule
                0    &  23.25     &  48.10  \\
                0.2  &  23.58    & \cellcolor{gray!20}\textbf{49.11}  \\
                0.4  &    23.25     &    48.57    \\
                0.6  &  \cellcolor{gray!20}\textbf{23.81}    &  49.08    \\
                0.8  & 22.78  &  48.54    \\ \bottomrule
            \end{tabular}
        }
        \caption{Different $\beta$ values}
        \label{tab:LossBalancing}
    \end{subtable}
    \caption{\textbf{Ablation Study on Pretraining Loss Design.} We present two sets of ablations: (a) a comparison of different loss configurations, evaluating CLIP-style versus CLCL supervision, and the impact of incorporating dual-level supervision (DS); and (b) the effect of hyperparameter $\beta$ in the final loss formulation (Eq.~\ref{eq:final loss}).}

    \label{tab:within_col}
\end{table}

\begin{table}[t]
\centering
\small
\begin{tabular}{l|l|cc}
\toprule
\textbf{Pretrained Setup} & \textbf{Init in Stage 2} & \textbf{BLEU4} & \textbf{ROUGE} \\
\midrule
None                        & None            & 16.01  & 39.23 \\
VLE                          & VLE              & 21.70  & 46.77 \\
\rowcolor{gray!20}\textbf{VLE + 3L TE}                  & \textbf{VLE}              & \textbf{23.81} & \textbf{49.08} \\
VLE + 12L TE                 & VLE              & 23.43  & 48.67 \\
VLE + 12L TE                 & VLE + TE         & 22.88  & 48.38 \\
\bottomrule
\end{tabular}
\caption{\textbf{Ablation of Pretraining Configurations.} We compare different encoder setups (\textbf{VLE}: Visual-Language Encoder, \textbf{TE}: Transformer Encoder with varying layers $L$) in pretraining and weight loading strategies for downstream translation.}
\label{tab:ablation}
\end{table}

\vspace{-1em}
\paragraph{Pretraining Paradigm.}
Finally, we examine how different pretraining configurations affect downstream translation performance, as shown in Table~\ref{tab:ablation}. Training the model end-to-end without any pretraining yields significantly lower performance (BLEU-4: 16.01, ROUGE-L: 39.23), highlighting the importance of visual-language alignment. In contrast, introducing a lightweight pretraining stage, where only the Visual Encoder and Language Mapper are trained using the Cross-Embedding Loss, leads to a notable improvement (BLEU-4: 21.70, ROUGE-L: 46.77), demonstrating that even shallow alignment provides strong benefits. Further adding a Transformer module during pretraining then enables dual supervision via both Cross-Embedding and Cross-Hidden-State losses. Interestingly, the 3-layer Transformer configuration outperforms its 12-layer counterpart, suggesting that deeper models may introduce unnecessary complexity and hinder alignment. Surprisingly, the best results are achieved when the 3-layer Transformer is used during pretraining but discarded during fine-tuning (BLEU-4: 23.81, ROUGE-L: 49.08). This highlights that while lightweight contextualization improves alignment, removing it downstream may enhance transferability by reducing overfitting to pretraining objectives.

\subsection{Qualitative Results}

\begin{figure}[t]
    \centering
    \includegraphics[width=1\linewidth]{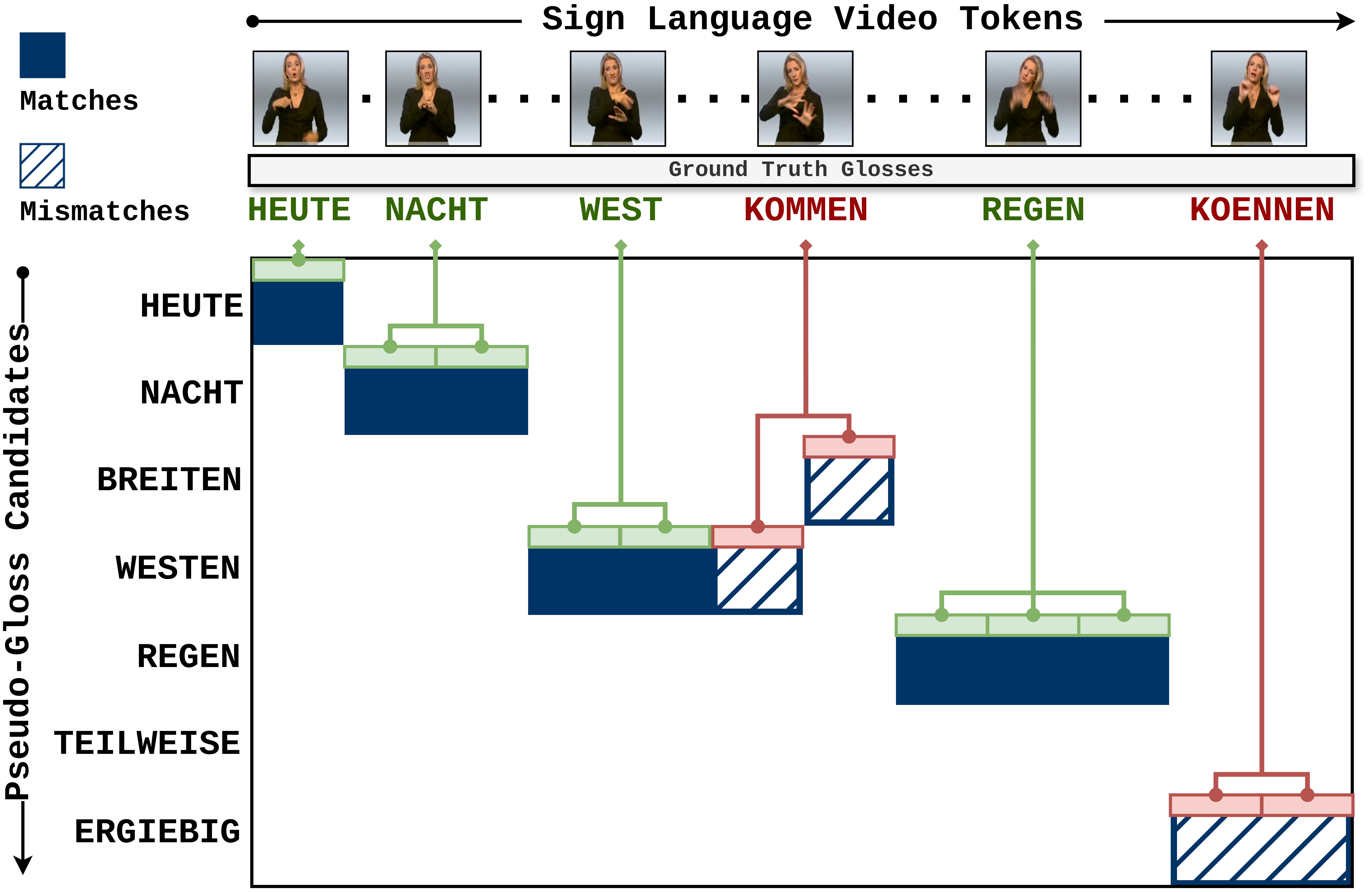}
    \caption{\textbf{Visual-to-Text Token Similarity Matrix.} The CLCL objective enables accurate fine-grained alignment between visual tokens and pseudo-gloss candidates. The solid blue boxes indicate correct matches, while the striped blue boxes mark mismatches.}
    \label{fig:CLCL}
\end{figure}

Fig.~\ref{fig:CLCL} shows the similarity matrix between segmented visual tokens and pseudo-gloss candidates via dot-product of their embeddings. The clear diagonal patterns show that the CLCL objective effectively enforces fine-grained, token-level alignment with semantically relevant pseudo-glosses.

However, since pseudo-glosses are derived from spoken language, they may not always accurately reflect the content of the signed video. For example, in Fig.~\ref{fig:CLCL}, terms such as \textit{Breiten}, \textit{Teilweise}, and \textit{Ergiebig} are present in the pseudo-gloss despite not being signed, while actual glosses like \textit{Kommen} and \textit{Koennen} are omitted. Such discrepancies can lead to occasional misalignments, as visual tokens may be left unmatched when no valid pseudo-gloss is present.

Despite these limitations, the pseudo-gloss representation still captures the majority of relevant glosses and facilitates effective alignment without requiring manually annotated gloss labels. This highlights the effectiveness of our framework, where segmentation-based tokenization and the CLCL objective jointly enables the learning of semantically meaningful alignments even under weak supervision. For further qualitative results, including additional alignment maps, spoken language predictions, and segment-level visualizations, please refer to the supplementary material.

\section{Conclusions}

In this work, we introduced a segment-aware tokenization framework for gloss-free SLT, leveraging sign segmentation for both localization and temporal downsampling. This reduces sequence length by 2× and memory usage by 2.67×, enabling more efficient and scalable training. We further proposed a dual-level alignment loss that enhances cross-modal supervision at both the input and language representation levels. On PHOENIX14T, our method surpasses prior state-of-the-art, especially under matched sequence lengths. These results highlight the effectiveness of our approach for efficient gloss-free SLT. Future work will focus on improving the visual encoders and scaling to larger corpora.

\footnotesize{
\paragraph{Acknowledgments:} 
This work was supported by the SNSF project ‘SMILE II’ (CRSII5 193686), the Innosuisse IICT Flagship (PFFS-21-47), EPSRC grant APP24554 (SignGPT-EP/Z535370/1) and through funding from Google.org via the AI for Global Goals scheme. This work reflects only the author’s views and the funders are not responsible for any use that may be made of the information it contains.}
{
    \small
    \bibliographystyle{ieeenat_fullname}
    \bibliography{main}
}

\newpage

\end{document}